\documentclass[a4paper,twoside]{article}

\usepackage{epsfig}
\usepackage{subfigure}
\usepackage{calc}
\usepackage{amssymb}
\usepackage{amstext}
\usepackage{amsmath}
\usepackage{amsthm}
\usepackage{multicol}
\usepackage{pslatex}
\usepackage{apalike}
\usepackage{csvsimple}
\usepackage{longtable}
\usepackage{float}
\usepackage{SCITEPRESS}     % Please add other packages that you may need BEFORE the SCITEPRESS.sty package.

\begin{document}

\title{Constructing a Word Similarity Graph from Vector based Word Representation for Named Entity Recognition}

\author{\authorname{Miguel Feria\sup{1}\sup{,}\sup{2}, Juan Paolo Balbin\sup{2} and Francis Michael Bautista\sup{2}}
\affiliation{\sup{1}Mathematics and Statistics Department, De La Salle University, Taft Avenue, Manila, Philippines}
\affiliation{\sup{2}Indigo Research, Katipunan Avenue, Quezon City, Philippines}
\email{\{juan\_feria\}@dlsu.edu.ph, \{miguel, jolo, francis\}@indigoresearch.xyz}
}

\keywords{named entity recognition, natural language processing, graphs, word networks, semantic networks, information extraction.}

\abstract{In this paper, we discuss a method for identifying a seed word that would best represent a class of named entities in a graphical representation of words and their similarities. Word networks, or word graphs, are representations of vectorized text where nodes are the words encountered in a corpus, and the weighted edges incident on the nodes represent how similar the words are to each other. We intend to build a bilingual word graph and identify seed words through community analysis that would be best used to segment a graph according to its named entities, therefore providing an unsupervised way of tagging named entities for a bilingual language base.}

\onecolumn \maketitle \normalsize \vfill

\section{\uppercase{Introduction}}
\label{sec:introduction}

\subsection{Named Entity Recognition}

\noindent Named Entity Recognition (NER) is a domain that is component to many natural language processing toolkits. NER is a method of information extraction that works to classify words in a collection according to their named entities. Examples of generic named entities include names of persons (PERSON), organizations (ORG), addresses or locations (LOCATION), measures of quantity, contact information, etc. Named entity types may be classified in a hierarchical manner, by which a location entity may have subtypes according to its purpose, like food locations, entertainment locations, religious locations, and the like \cite{Palshikar}.

There are multiple approaches for NER classification for words. Rule-based methods are typically constructed by domain experts in syntax and language, and are handcrafted and manually built. Supervised learning approaches make use of pre-existing corpora of hand-tagged named entities, and algorithms are used to identify the latent rules necessary to classify the named entities.  Unsupervised approaches to using machine learning to classify NERs involve using seed words of the named entity class by which to group a collection of words against. The score for grouping membership determines the words NE class and allows the model to continue determining new rules should the word collection be improved.

The task of automating the process of named entity recognition presents multiple challenges. One approach to NER is to maintain a list of named entities by which a system will check against when analysing a text. A library of reference words and their variants would have mappings to their pre-classified named entity types, and any new word to be classified would have to be referenced against the pre-classified library. This presents a dependency to the comprehensiveness of the pre-classifications and how frequently this is updated.

Capitalisation of words also contributes to the problem, as words may have multiple meanings based on their capitalisation and placement in a sentence. Certain words may have different meanings based on the context of its usage, and capitalization adds complexity to this problem by expanding the search space for a named entity. \cite{Palshikar}

Another problem arises when we consider NERs that are made up of multiple words, as boundaries for NEs are difficult to determine. Two unconnected words may have different named entities, but when used as a compound word, may have an entirely different named entity classification. \cite{ner_fil}

In the case of this paper, the bilingual nature of our corpus poses additional complications. For a bilingual model, NE groups should contain words from both English and Filipino. Potential overlaps between NE groups then may arise, and determination of the correct NE may be more complex. 

\subsection{Graph representation of NER}

Identifying seed words to build NER classification types on a dataset would provide opportunities for generic NER models that would work on a corpus. In particular, in this paper we will be using a billingual corpus for NER. The unsupervised approach to building a multilingual word graph and using seed words to tag NEs would reduce the dependency on manually tagged NLP libraries. While representing bilingual word similarity through a graph data structure presents multiple challenges, this would open research for NE boundary identification, and understanding how similar or dissimilar languages are to each other by analyzing them graphically.

\paragraph*{Vector Representation of Words}

One way to quantitatively evaluate the similarity between two words is to apply the vector representation of words. \cite{Mikolov_2013}, proposes a method for representing words into vector space using statistical techniques, resulting in lower computational complexity than previous LSA and LDA models (citation). This method results in an n-dimensional vector representation of a word, from which the similarity of a word to another can be computed using simple linear matrix operations. A vector space is then constructed such that word embeddings of words that may belong to a similar context within a sentence are close together in the space. 

Our word embedding model was trained over a bilingual corpus, as such, we will be able to determine the similarity between English and Filipino words. This presents an opportunity to observe semantic relationships between words in from different languages. We hypothesize that community structures in the graph representation of our model will reveal these semantic relationships, and facilitate NER for a bilingual corpus. 

\subsection{Related Work}

There has been prior work in determining named-entities using graphical models to identify the NE boundaries. 

\cite{Palshikar}, in his work detailing NER identification techniques, discussed unsupervised methods in identifying named entities. One of the disadvantages of supervised approaches is usually having to source and work with large labelled datasets. These labelled datasets are typically manually tagged by linguistic experts that have to agree on which rules to adhere to, and consume a lot of time to develop. 

Unsupervised methods to identifying NEs work on untagged datasets by beginning with a small set of NE tagged seed words. The algorithm would identify words that are similar in position, occurrence and co-occurrence, and develop a boundary to separate one NE from others. Once the model is generalized, it can be used on other bodies of text to identify NEs. Palshakir 2012 argues that the unsupervised method to identifying NEs is more aligned with how language typically evolves; bootstrapped unsupervised NE recognition can adjust to the varying tolerances of how language changes over time.

The work done by \cite{yhu2015} in identifying unsupervised approaches for constructing word similarity networks utilized vectorized corpora as a base dataset. They discussed the use of Princeton’s WordNet as a template for a word association network as their motivation. \cite{yhu2015} contributed to this model by measuring the information of a word’s co-occurrence with another word. Once the vectorized representation of the word’s cosine similarity score with another word is computed, they construct a similarity network where the weighted edges between words are selected above a specified threshold.

In work done by \cite{Hulpus_2013} researchers created sense graphs $G=(V,E,C)$, for  topic $C$, such that $C \in V$ is a seed concept or seed word, to model topics in DBpedia. They constructed a DBpedia graph, and using centrality measures, were able to show which words best represent topics in DBpedia. The researchers suggest several centrality measures to identify the best seed words, and present their experimental results for several focused centrality measures. 

\cite{Hakimov_2012} also employed graph based methods to extract named entities, and disambiguation from DBpedia. Their methodology also employed graph centrality scores to determine the relative importance of a node to its neighbouring nodes.To determine word similarity, they spotted surface forms in a text and mapped these to link data entities. From these relationships, they constructed a directed graph, and disambiguated words. They compared their methodology to two other publicly available NER systems and showed that theirs performed better. 

\section{\uppercase{Methodology}}

\paragraph*{Corpus}

The corpus used for this study is comprised of comment data from 21 public Facebook pages extracted using the Facebook public API. The comments came from posts spanning January 2017 to October 2017. The length of the each comment in the dataset varies, and each comment was stripped of emoji and other noise characters and metadata like stickers and images. In total, the corpus contained around 7.5 million comments with around 222,000 unique words and is at 375 mb in size.

The corpus includes a bilingual dataset covering English and Filipino words, and their combination. The stop words like \textit{it} and \textit{so} were not removed to preserve the statistical connections between each word. Stemming wasn't also done for the words included in the corpus.

\paragraph*{Word Embedding}

The first step in our methodology involves creating a word embedding model that maps words to a vector space. The method involves a two layer neural network model that was trained on the comment dataset corpus. The model produced a word matrix where each word in the corpus is mapped to a vector in the matrix.\cite{Mikolov_2013}

There are two common architectures used to model word embeddings: the continuous bag-of-words (CBOW) method, and the continuous skip-gram method. Our model employs the CBOW architecture, as such,  our study is scoped by the parameters of this method. \cite{Mikolov_2013} Future studies can improve on this methodology by comparing results with the skip-gram method of training a word similarity model. In the CBOW architecture, the order of words in a sentence is not considered, thereby not affecting word similarity measures. 

The result of the training over the dataset is a model that quantifies the similarity between words in the model's vocabulary $V_{m}$. Given a word $w_{i} \in V_{m}$ it is related to another word $w_{j}$ with similarity score $w_{ij}$, where $0 < w_{ij} <1 $. 

\paragraph*{Constructing the Graph}

We use the output of the word embedding model to create a graph representation for the similarity between the words in our model’s vocabulary. In our analysis, we take each word from the vocabulary and apply the word embedding model to obtain a list of words similar to it, along with the similarity score. We then construct a graph $G = (W, S) $ where $W$ is the set of words ${w_1, w_2, \ldots, w_{|W|}}$, and $S$ is the set of edges $S = \left\{(w_i,w_j, w_{ij}) | w_{i,j} \in W\right\}$ connecting words with their corresponding similarity score as weight. 

A characteristic of the word embedding model is that it produces similarity scores for every word pair, that is, there is a relationship and a similarity score between every word in the vocabulary. To avoid producing a complete graph, we set a limit for the number of similar words, as well as a lower bound for the similarity score. In our experiments we set these limits with the intention of reducing the number of edges, while still preserving the overall structure of the graph. 

\paragraph*{Community Detection}

Our approach to determine the named entity clusters from the word-network we obtained is to subdivide the graph into smaller graphs consisting of highly interconnected nodes. Literature suggests that community detection in information networks may help uncover topics within the information network. \cite{Fortunato2012} In this paper we will use the Louvain Algorithm for community detection. The Louvain Algorithm uses an iterative process that optimizes the modularity of the graph. Modularity is a value between -1 and 1 that measures the ratio between edges inside communities and the edges that connect the communities. It is given by the following equation: 
$$Q = \frac{1}{2m} \sum_{i,j}[w_{ij}-\frac{k_{i}k_{j}}{2m}]\delta(c_i,c_j) $$

Where  $k_i = \sum_{j}w_{ij}$ is the sum of weights of the edges adjacent to $w_i$, $c_i$ is the community $w_i$ belongs to, $m = \frac{1}{2}\sum_{ij}w_{ij}$, and the function $\delta(c_i,c_j)$ has a value of 1 if $c_i = c_j$ and 0 otherwise. It is clear that the value of Q goes to zero when $w_{ij} = \frac{k_{i}k_{j}}{2m} $, that is, when we consider the entire graph as one community. 

The proposed algorithm works by initially assigning each node of the graph to a unique community, and then for each node $w_i$ in each community, consider the nodes $w_j$ adjacent to it. For each node $w_j$, the adjacent node $w_i$ is added to $w_j$'s community and modularity is computed. 

For all nodes $w_i$ adjacent to $w_j$, the algorithm computes for the modularity score and retains the node that returns the largest increase in computed modularity score. This is done iteratively as the communities are built, and stops when there is no increase in community modularity when node $w_i$ is added. \cite{louvain}

\paragraph*{Betweenness Centrality}

Betweenness centrality is a centrality measure in a graph that considers the number of shortest paths a node lies on. 

In network analysis, we may consider nodes as more important than others. Generally, centrality scores are used for this. Betweenness centrality measures the ease of flow of information between other nodes in the graph, or community. Our hypothesis is that nodes with high betweenness centrality are ideal seed words for NER. We will compute for betweenness centrality of every word in each community, and return the largest centrality value, which should give us the word that is most similar to the other words in the community it lies in. This will be of great use for NER, as seed words can contain rich lexical information on the relationship between the words in a community. 

For a weighted network, betweenness centrality is computed using the following formula: 
$$ C_b (v) = \sum_{w_i \ne w_j \ne w_k \in V} \frac{\sigma_{ij}(v)}{\sigma_{ij}} $$ 

where $\sigma_{ij}$ is the length of the shortest path from nodes $w_i$ to $w_j$, and $\sigma_{ij}(v)$ is the length of the shortest path from nodes $w_i$ to $w_j$, passing through node $v$. \cite{newman2010networks} Weighted path lengths are computed by summing up the weights of the edges traversed by the path. \cite{Brandes_2001}

\section{\uppercase{Evaluation and Results}}

All the data analysis is done through Python. The \textit{Gensim} package is used to implement \textit{word2vec}, the vector space mapping algorithm. \textit{NetworkX} is used for the basic network analysis and computation of betweenness centrality for the graphs constructed. \textit{Python-louvain} is a community detection package that works alongside \textit{NetworkX}. It implements the Louvain algorithm for optimal modularity. 

Our resulting graph was made up of 222,239 nodes, and 9,766,691 edges. We were then able to identify 26 communities using the Louvain Modularity algorithm.

We believe that due to the nature of our dataset, the precision of our results could be improved on. The dataset contains a lot of comments that contain spelling and grammatical errors in both English and in Filipino. However we were able to extrapolate some contextual information from certain communities. 
\subsection{Community Detection}
\begin{figure}
\begin{center}
\includegraphics[scale=0.25]{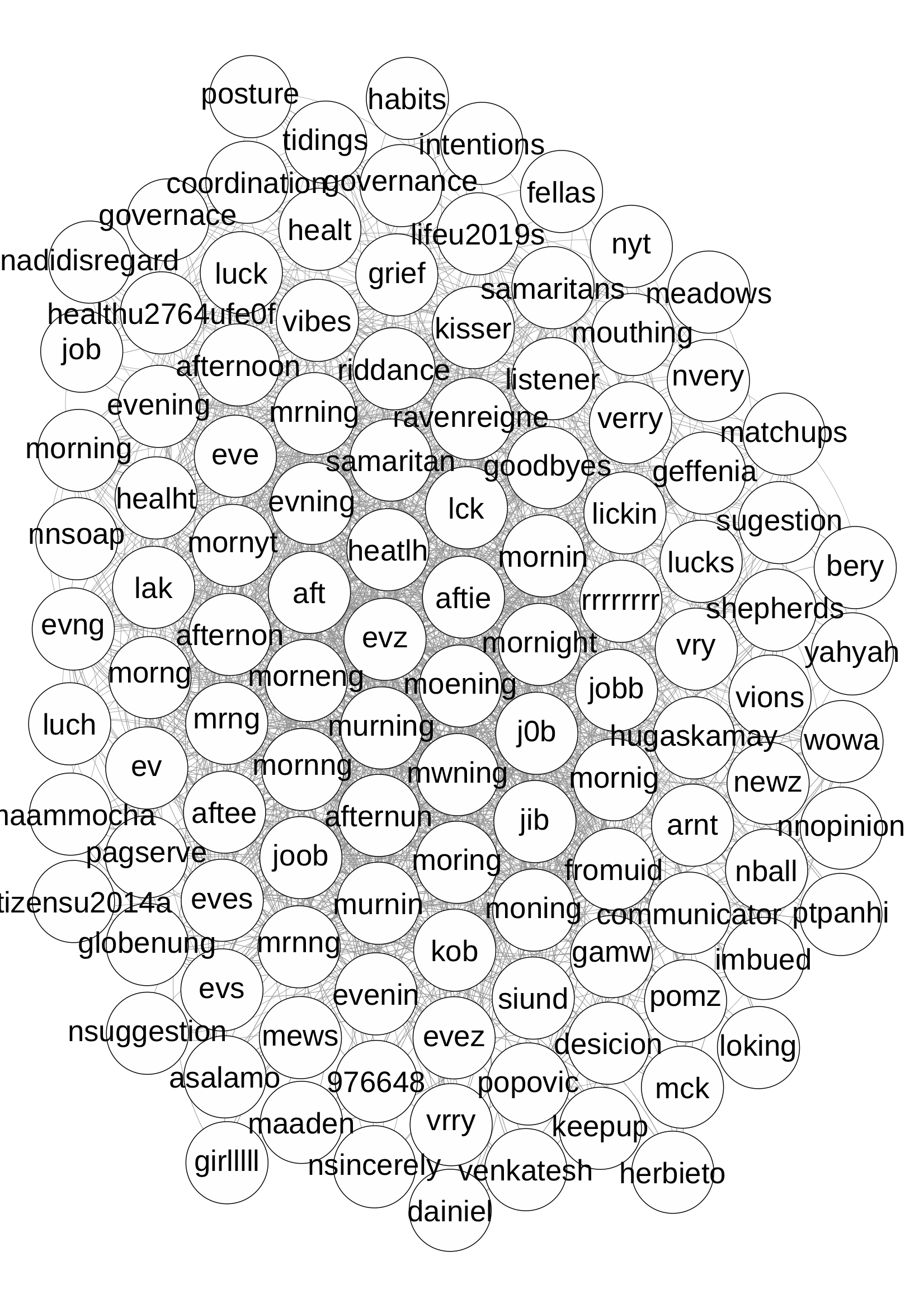}
\caption{Community of words that are related to the word \textit{good}, as in \textit{good j0b, good morning, good riddance, good vibes, verry good, lickin good}}.
\label{good_comm}
\end{center}
\end{figure}

From one community, we found that words that are often used in conjunction with \textit{good} were grouped together. Words such as \textit{morning, eve, health, riddance, samaritans} and \textit{afternoon} were found to be very close to each other in the graph. Words that were misspelled also made it to the graph, as seen in figure \ref{good_comm}. Other forms of \textit{good} are also present in the corpus such as \textit{gud, gd}.

Another such community we found was one that revealed words that are often paired with \text{happy} as in most greetings. Words such as \textit{birthday, anniversary, valentines}, and ordinal numbers associated with these greetings, such as \textit{1st, 50th, 52nd} and other misspelled numbers and words. 

\begin{figure}
\begin{center}
\includegraphics[scale=0.8]{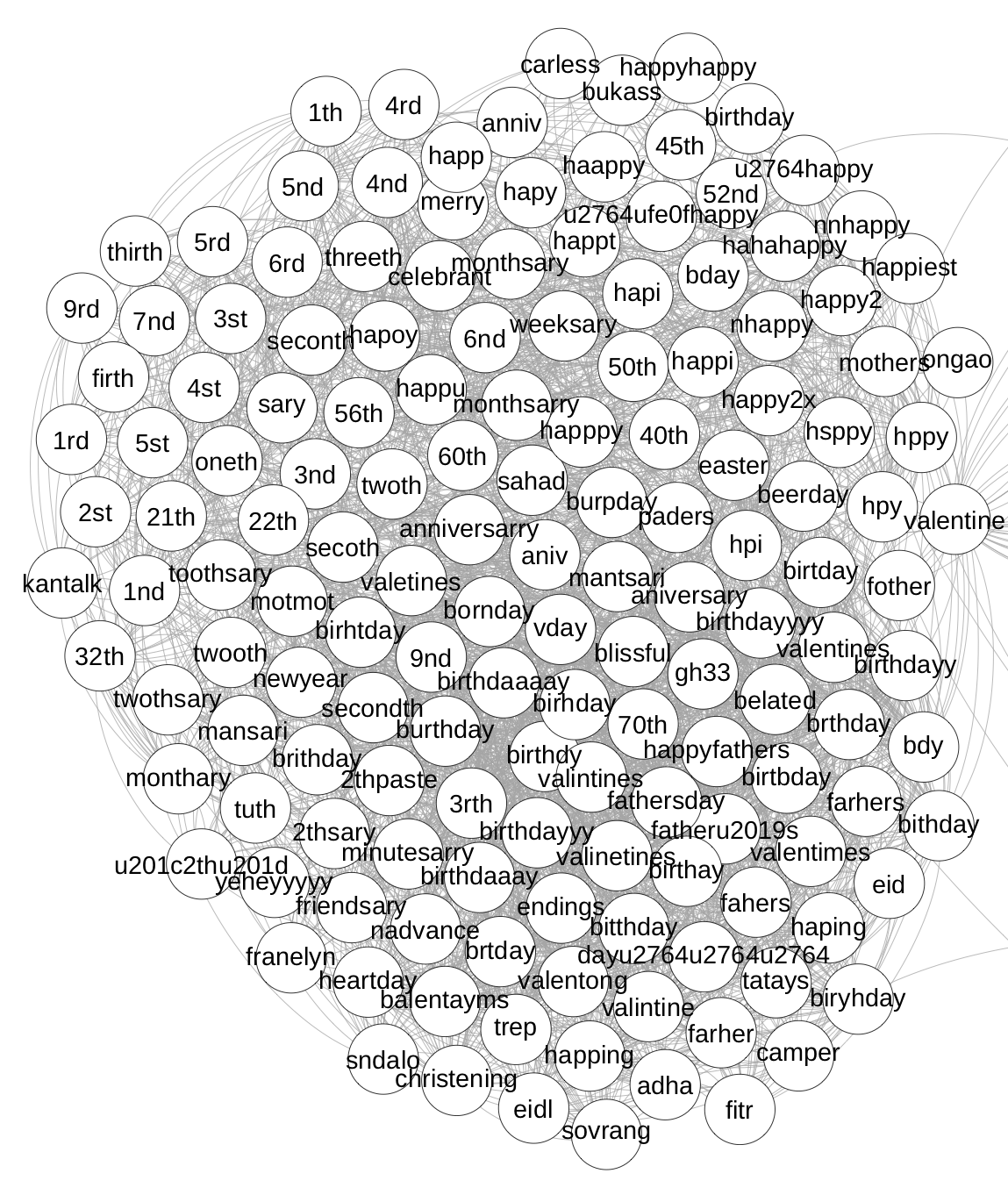}
\caption{Community of words that can act as greetings, and ordinal numbers associated with the greetings.}
\label{greetings_comm}
\end{center}
\end{figure}

\subsection{Seed words}

For the community in figure \ref{greetings_comm}, we found that the node with highest betweenness centrality is \textit{valentine}, with centrality score \textit{0.1833732}. This validates our hypothesis that words in this community are words that often appear next to the word \textit{happy}, and as such the community can be considered a NE for greetings, composed of special occasions, and ordinal numbers. For the community found in figure \ref{good_comm} the most central node was found to be the word \textit{fromuid}, with centrality score \textit{0.1187109}. The next most central nodes yield more lexical context for our analysis: \textit{jib, moning,} and \textit{moring}, with centrality score \textit{0.1080807, 0.0896776, 0.0861839}, respectively. We believe that these words are misspelled versions of the words \textit{job} and \textit{morning}, which as we hypothesized earlier, are words that are commonly used with \textit{good}. 

Another community that we discovered was a community of Filipino words that work as a means of expressing adjectives to a superlative degree. The word \textit{kaingay}, which is a shorted version of \textit{napakaingay}, is a close translation of to be very noisy---\textit{ingay} translates to noisy, and the prefix \textit{ka-} maps to a higher degree of what it modifies. The community includes words like \textit{kalake}, \textit{kapalpak}, \textit{kaluwag}, and \text{katagal} that fit the following classification of Filipino words that are all prefixed with \textit{ka-} as a modifier for degree for an adjective.  

\section{\uppercase{Conclusions}}
\label{sec:conclusion}

\noindent In this paper we proposed a methodology to obtain named entity groups from a vectorized vocabulary of words, by constructing a word similarity graph, and observing the communities found in the graph. We applied this methodology to a bilingual dataset mined from around 300 megabytes of public Facebook comments and found 26 communities. The named entity groups we found were not what had sought out to find at the beginning of this project, but we were able to find that community structures in graph reveal more than just basic named entity groups such as names of persons (PERSON), organizations (ORG), addresses or locations (LOCATION), or measures of quantity. These community structures may reveal deeper lexical concepts, such as \textit{greetings}, \textit{superlative degree} or \textit{things that are good}. The revelation of additional named entity groups could provide greater insight on how language construction, at least in the online fora, is statistically used.

Future work could also involve applying this methodology to a cleaner dataset, that is, a dataset with stricter grammatical restrictions. This could provide the development of stronger communities that may provide other classifications of NEs. This method of identifying NEs may augment existing NER methods and may benefit documents and datasets that may be more domain specific like \textit{medical documents} or \textit{legal proceedings}.

\noindent Other centrality measures could also be investigated. Eigenvector centrality variants like PageRank or the Katz centrality could be used to identify the most influential word in a community and provide lexical information on seed word identification. \cite{newman2010networks} This could be beneficial in automatically extracting seed words in a community without having to inspect each community individually.

\noindent As a more macroscopic recommendation, exploration could be done to see how English words and their Filipino counterparts relate to one another graphically. An analysis following this approach might reveal more lexically similar words across the bilingual word space. Conversely, it might also present words and or concepts that are more unique to a specific language base. The identification of these graphical constructs across a bilingual language base might aid machine translation efforts.
~
\end{document}